\documentclass[10pt]{bmc_article}

\usepackage{cite} 
\usepackage{url}  
\usepackage{ifthen}  
\usepackage{multicol}   
\usepackage{amssymb}
\usepackage{amsmath}
\usepackage{graphicx}

\usepackage[utf8]{inputenc} 
\newboolean{publ}

\newenvironment{bmcformat}{\begin{raggedright}\baselineskip20pt\sloppy\setboolean{publ}{false}}{\end{raggedright}\baselineskip20pt\sloppy}

\begin{document}
\begin{bmcformat}


\title{Automatic ECG Beat Arrhythmia Detection}
 

\author{M. Bazarghan\correspondingauthor$^{1}$%
       \email{Mahdi Bazarghan\correspondingauthor - bazargan@znu.ac.ir}%
      \and
	Y. Jaberi $^2$%
         \email{Yahya Jaberi - yahya.jaberi@zums.ac.ir}%
      \and
        R. Amandi $^3$%
         \email{Ruhallah Amandi - amandi@azu.ac.ir}%
      \and
        M. Abedi $^4$%
         \email{Majid Abedi - abedi@iasbs.ac.ir}%
      }


\address{%
    \iid(1)Department of Physics, Zanjan University, Zanjan 313, Iran\\
    \iid(2)Zanjan University of Medical Sciences, Zanjan, Iran\\
    \iid(3)Department of Computer Engineering, Islamic Azad Univeristy, Zanjan Branch, Zanjan, Iran \\
    \iid(4)Institute for Advance Studies in Basic Sciences, Zanjan, Iran
}%

\maketitle


\begin{abstract}
        \paragraph*{Background:} 
        In  recent years  automated data analysis techniques have drawn great attention and are used in almost every field of research including biomedical. Artificial Neural Networks (ANNs) are one of the Computer- Aided- Diagnosis tools which are used extensively by advances in  computer hardware technology. The application of these techniques for  disease diagnosis has made great progress and is widely used by  physicians. An Electrocardiogram carries vital information about  heart activity and  physicians use this signal for  cardiac disease diagnosis which was the great motivation towards our study.
      
       \paragraph*{Methods:} 
       In this study we are using Probabilistic Neural Networks (PNN) as an automatic technique for  ECG signal analysis along with a Genetic Algorithm (GA).  As every real signal recorded by the equipment can have different artifacts, we need to do some preprocessing steps before feeding it to the ANN. Wavelet transform is used for extracting the morphological parameters and median filter for data reduction of the ECG signal. The subset of morphological parameters are chosen and optimized using GA. We had two approaches in our investigation, the first one uses the whole signal with 289 normalized and de-noised data points as input to the ANN. In the second approach after applying all the preprocessing steps the signal is reduced to 29 data points and also their important parameters extracted to form the ANN input with 35 data points.

        \paragraph*{Results:} 
        The outcome of the two approaches for 8 types of arrhythmia shows that the second approach is superior than the first one with an average accuracy of \%99.42 . 

        \paragraph*{Conclusions:}
        We have studied 8 types of arrhythmia with high detection accuracy. In the literature, previous attempts are made on 6 types of arrhythmias. The results of the PNN shows that its performance for reduced input signal along with the morphological parameter has the best performance. Also it was noticed that the proper selection of training and testing data sets are of great importance and all the beats of each arrhythmia should not be selected from a single file in the database.
\end{abstract}

\ifthenelse{\boolean{publ}}{\begin{multicols}{2}}{}


\section*{Background}
 The electrocardiogram (ECG) inherently carries important information on functionality of the heart. This signal provides a physician with crucial information on a patient's heart function and can be used for the prognosis and diagnosis of  heart disease. It is one of the most common signals used in diagnosis because of its non-invasive nature and the valuable information it contains. Its analysis can be used to judge the pathophysiological condition of the heart. Several systems have been developed for ECG recording and analysis. Early ECG systems were just recording the signal by printing it. New systems use computer technology to provide automated diagnosis. The latter is a large research field and many methods and approaches have been proposed and implemented for the detection of ischemia, arrhythmia detection and classification, and diagnosis of chronic myocardial diseases. Those methods usually include  processing of the signal and removing noise and artifacts, extracting certain key features related to diseases, and analyzing the features to make the final decision. The analysis is usually done by using signal processing, artificial neural networks, and fuzzy logic concepts along with the clinical symptoms provided by medical experts. The performance of those systems is evaluated using standard databases.\\ 
The normal activity of the heart starts at a region in the upper wall of the right atrium known as the Sino Atrial (SA) node or the heart’s pacemaker, and through both atria forwards to the Atrioventricular Node (AV) and enters the ventricle from the AV node. There are different types of heart rhythm disorders or arrhythmias. In cardiac arrhythmias,  the electrical activity usually starts at a location other than the sinus node or the propagation and its speed is abnormal. Tachycardia and ventricular fibrillation are  examples of dangerous arrhythmias. Atrial premature beats are other examples of cardiac arrhythmia.\\

The mechanical and electrical activity of the heart is periodic. The electrical activity of the heart at each cycle generates  recordable potential differences of, 300 to 1000 ms during which the voltage as well as the direction changes rapidly. These electrical vectors can be recorded by fixing the electrodes on the patients' skin. The ECG record on the skin in one direction will not give enough information because generated electrical vectors by the heart are changing and moving in three dimensional space. That is why ECG has 12 vectors recording signals in 12 dimensional space where 6 of them are in the horizontal plane and the other 6 in the frontal plane; they are all called ECG leads. These leads record signal by placing electrodes on the patient's upper and lower limbs and chest. A physician analyses the ECG by having prior knowledge on the space angle of each lead and looking at the plotted signal of the leads.\\

A normal ECG signal includes a set of waves and each one represents the activation or inactivation of certain anatomical parts of the heart. These waves are called P, QRS, and T waves which are shown in Figure \ref{nsr}. The P wave is the result of activation of the atria and the QRS wave by activation of the ventricles. As the activation of QRS wave is a complex process, so the QRS wave will also be complex. The ventricle inactivation produces the T wave. A wave corresponding to atrial inactivation occurs, but is ordinarily buried in the QRS complex, and is not identifiable in the ECG. Arrhythmia detection and some  heart disease diagnosis is done by ECG analysis and is usually done by the heart specialist.\\

\begin{figure}[ht!]
\centering
\includegraphics[width=8cm]{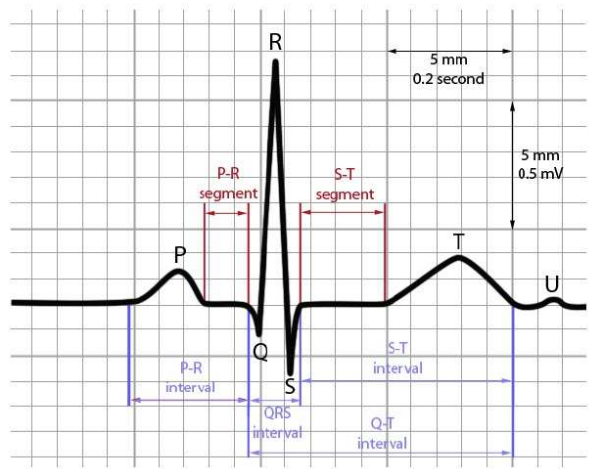}
\caption{The sample normal beat showing different morphological parameters.}
\label{nsr}
\end{figure}

Over the last few decades with the entering of computers into our daily lives,
they've come to play more  important roles in medical science. One of the main
areas that computers entered in medicine is disease diagnosis. The Computer
Aided Diagnosis (CAD) has made great progress in the last two decades
\cite{Yang,Kunio,Steward,Sutton}. Computers have drawn the attention of
physicians and engineers in the areas like Lung Cancer \cite{Lung}, lung
disease \cite{Lung1,Lung2,Lung3}, breast cancer \cite{Breast} Colonography and
polyp \cite{Polyp1,Polyp2}. The first attempts to use computers in medicine
for automatic diagnosis started in the 70s, and later in the 90s with further
studies led to CAD. The computer processing results used for disease diagnosis
are nowadays not used in place of physicians but beside them; meaning, the physician gives the final diagnosis with the help of physician assistant tools. One of the important tools in the CAD is Artificial Neural Networks. The ANN is a system loosely modeled based on the human brain and has wide applications in almost every field of science including medical science. ANNs with the ability to recognize patterns and processing large amount of data can help physicians in disease diagnosis \cite{IEEEmed}. Nowadays, ANNs are widely used in the diagnosis of, Ovarian cancer \cite{Ovarian}, Prostate Cancer \cite{Prostate}, Pancreatic disease \cite{Pancreatic}, EEG abnormality detection \cite{EEG}, ECG signal \cite{ECG}. The number of published papers on applications of ANNs in medicine has increasingly risen in the recent years \cite{Brause,Annrev1,Annrev2}. \\

The ANN uses non-linear mathematical models to solve problems.  As humans use prior information to solve new problems, ANNs also use solved examples to construct neuron systems to recognize new patterns \cite{eftekhari}. In recent years with the advances in computer processing capabilities, ANNs have drawn great attention in different areas such as; space science, transportation, gas and petroleum, and medicine. \\

The analysis described here uses ANNs for ECG signal analysis and cardiac
arrhythmia diagnosis. The methods section describes the data and the
abnormalities present in them. It gives the preprocessing steps and explains
the parameter extraction of the data and give the classification methods. The
last sections give the results and conclusions.

\section*{Methods}

\subsection*{Data}
There are different databanks of cardiac signals. In this work we have used
the MIT-BIH database \cite{MIT-BIH}. This databank consists of 48
signalsrecords of 30 minutes from 48 different patients. The sampling
frequency of this data is 360Hz and the sampling time rate is
2.7ms. Approximately 60 percent of these records are for patients with
arrhythmia.  25 of 48 records are less common but clinically significant
arrhythmias. These data are recorded by Holter \cite{Gao}. In this work we use
the data recorded by lead 2. Each 30 minute long record is a continuous
waveform and comes with annotations for each heartbeat. These annotations
represent the type of arrhythmia. Since we were interested in only certain
types of arrhythmias in the database, atfor each of the 48 records, we extracted interesting individual beats with their associated annotations. The list of arrhythmias which we have investigated in this work with their annotations are given in Table1. \\

A total of 2800 beats were extracted with 350 beats belonging to each category viz. Normal, Paced, Premature, Escape, Fusion of paced and normal, Fusion of ventricular and normal, Right Bundle Branch Block and Left Bundle Branch Block. Six different datasets were formed from the complete data. The sample plot of each eight arrhythmias under our consideration is given in Figure \ref{arryt}.

\begin{figure}[ht!]
\centering
\includegraphics[width=16cm]{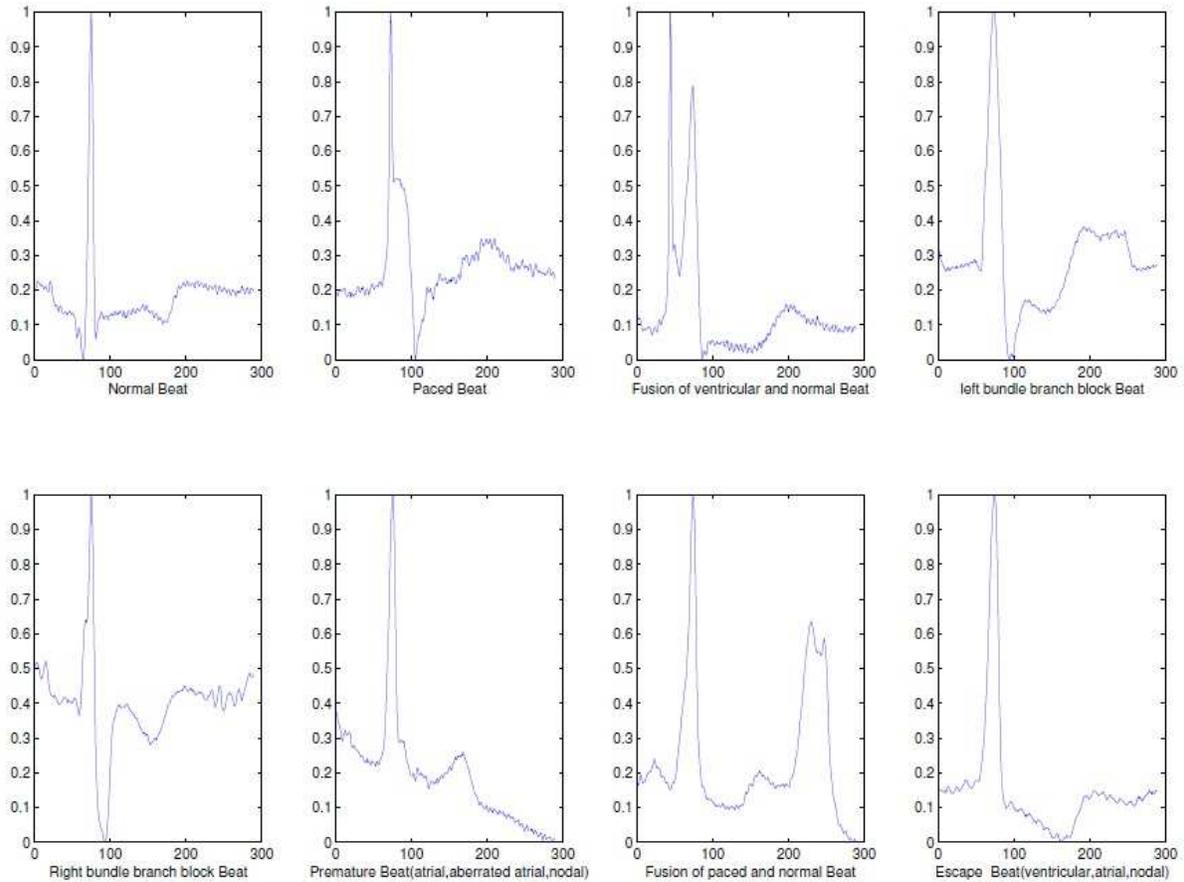}
\caption{Sample signals for eight arrhythmias under consideration.}
\label{arryt}
\end{figure}


\subsection*{Pre-processing}
The analysis of these data requires some preprocessing steps. The performance of the algorithms applied to the data is very much dependent on these preprocessing steps which are explained below. 
\subsubsection{Beat separation}
The first step in the preprocessing of signals in this application is to
extract the beats for classification of the normal and abnormal beats. For
extracting the beats we need to first find the peak of the ‘R’ wave then take
200ms before and 600ms after this peak value as a complete beat. According to
practitioners the information for arrhythmia detection lies in this 800ms
period, and other parts of the signal do not carry significant
information. This 800ms signal usually consists of the entire P, QRS, and T waves.
\subsubsection{Normalization}
Normalization is one of the important preprocessing steps, and an ANN with
normalized input vectors gives better results and converges faster. Here we
have used the Min-Max normalization method.  In this method a linear mapping is done, in which, the maximum value among vector elements is mapped to 1 and the minimum is mapped to 0. In other words, the components of normalized vector ($v'_i$) can be found as:
\[V'_i=\frac{V_i-min}{max-min}\]
where, $V_i$s are the components of the input vector. And min and max are the
minimum and maximum value of the components of the vector respectively\cite{Dmandkd}.
\subsubsection{Data reduction}
If the sampling time steps are small, then the size of each heart beat data
will be large. The larger the data size, the larger the memory required and
the longer the time for processing it. So, it is desired to reduce the size of
the data as much as possible. For which, the available filtering methods can
be used, and we are using median filter. The median filter is actually
replacing the number of consecutive data points with their median value. We
are using each 10 consecutive data points of signal and replacing them with
their median value to reduce the data size and is shown in Figure \ref{median}. In this way our input vectors with
289 data points reduce to vectors with only 29 data points. This leads to not
only data size reduction but also reduces the signal’s noise, because, by applying this filtering method the instant noise present in the signal vanishes. Furthermore, data size reduction will also reduce the complexity of the ANNs and hence its confusion, which can lead to faster convergence of the ANNs.

\begin{figure}[ht!]
\centering
\includegraphics[width=16cm]{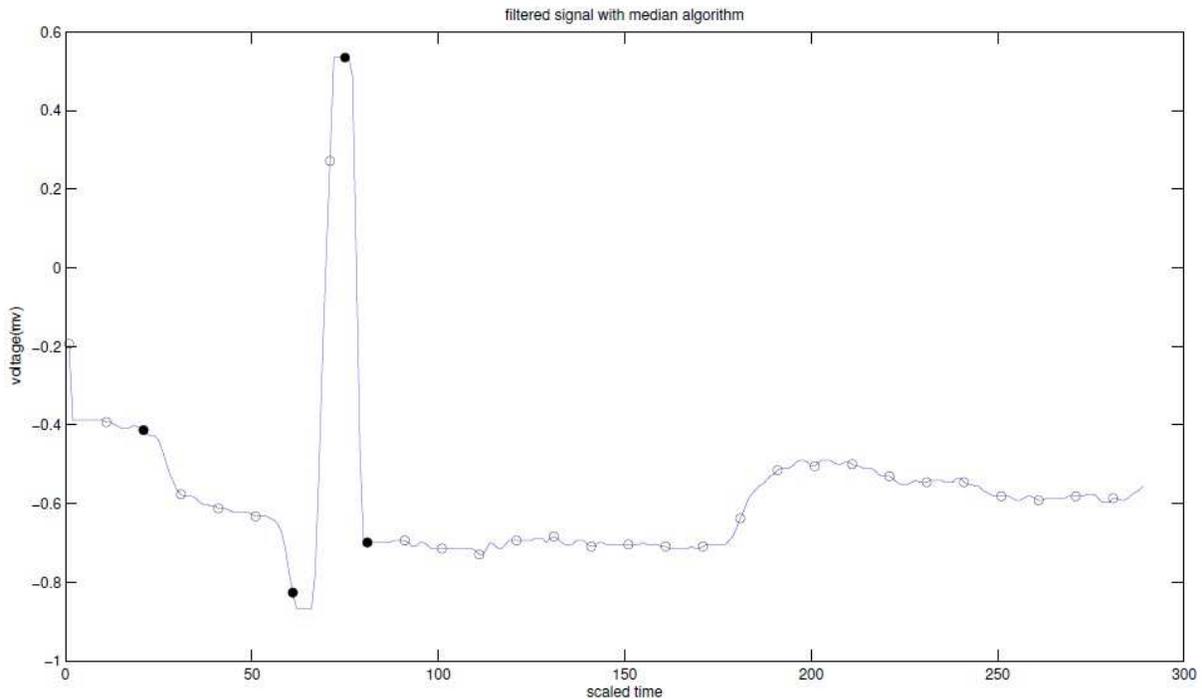}
\caption{Median algorithm used for data reduction and 10 consecutive points are
  replaced by a single point with their mean value.}
\label{median}
\end{figure}

\subsubsection{De-noising}
ECG signals are recorded using surface Electromyography(EMG).  Voltage sensors
are placed on the skin in the vicinity of the heart. The signal measurement
and the initial signal processing with the instrument creates some noise. The
presence of noise in ECG signals may cause problems in identifying the arrhythmia.
\\
The five main sources of noise are, power line interference, EMG noise, electrode contact noise, motion artifacts and instrumentation noise. Usually filters are used to remove unwanted noise artifacts. 
\\
The power line interference primarily consists of two mechanisms: capacitive
and inductive coupling. Capacitive coupling is due to an undesired capacitor
formed between two elements of a circuit in which the energy transfer occurs. Inductive coupling on the other hand is caused by mutual inductance between two conductors of the elements.
\\
Electrode contact noise is caused by changes in the position of the electrodes on the skin and also the medium between the heart and the electrodes.  These changes affect the impedance, meanwhile the impedance in the circuits is assumed to be constant to measure the voltages. If the electrode-skin impedance is large enough, the relative impedance change cannot shift the baseline of the ECG signal, but in practice it is not.
\\
The position with respect to the heart should be constant but movements of the heart and also the electrode on the skin cause motion artifacts. As the movements are slow these artifacts cause peaks in the Fourier power spectrum in low frequencies compared to the frequency of the heart rate. 
\\
EMG noise is due to the contraction of other muscles besides the heart. This contraction generates depolarization and repolarization waves near the electrodes that is picked up and summed into the main waves of the heart signal. 
\\
The presence of noise in all electrical elements causes the Instrumentation noise. The electrode probes, cables and amplifier are the primary elements for ECG. These three elements and any other element like Analog-to-Digital converter and Signal processor contain white Gaussian noises themselves. 
\\
These noise in the ECG signal are to be removed for the automatic disease diagnosis. In most of the signal processing cases, de-noising is one of the important and required steps for proper signal processing. Here we used Wavelet transform for de-noising\cite{Waveletdenoise} and the first order Coiflet from Matlab toolbox is used for this transformation which is explained more in detail in the next section.



\subsection*{ECG parameters}
In addition to the recorded signal and using it as input to the ANN, the
important extracted features from the signal can also be concatenated. These
features, which are shown in Figure \ref{nsr}, play a key role in disease
diagnosis. Each single bit can be characterized with the following parameters;
QRS duration, QRS amplitude, RR interval, PR segment, PR interval, P
amplitude, ST segment, ST interval, QT interval, T amplitude. In fact the type
of abnormality and cardiac arrhythmia directly influences these parameters and
the duration or amplitude of each wave will be affected correspondingly.

\subsubsection{ Feature set selection using Genetic Algorithm}
The main requirement for diagnosis systems is to achieve high prediction accuracy. Furthermore, a classification learning algorithm is expected to have short training and prediction times. 
In real classification problems one needs to choose a subset of features from
a much larger set that represents the knowledge to be used in the
classification. This is because of the fact that the performance of classifier
and the cost of classification are sensitive to the selection of the features
used in the construction of the classifier. By reducing the set of features, the time required for learning the classification knowledge and the time needed for classification reduces \cite{Featurega}.
Further, by the extraction of relevant features and therefore the elimination
of the irrelevant ones, the accuracy of the classifier can be
increased. Exhaustive evaluation of possible feature subsets is usually
infeasible and not recommended in practice because it requires large amounts
of computational effort. This is where the Genetic Algorithms (GAs) can be
very handy and offer an attractive approach to find near-optimal solutions to
such optimization problems. GAs are adaptive heuristic search used to solve
optimization problems guided by the principles of evolution and natural
genetics \cite{Clever}. In GAs, the parameters of the search space (in our
example, 10 ECG morphological parameters) are encoded in the form of strings,
called chromosomes. A collection of such strings is called a population. In
the case of feature selection problem, each chromosome represents a subset of
features selected. The number of these features defines the size of a
chromosome. Each element of the chromosome string can take the value 1 or 0,
where 1 indicates that the corresponding feature is selected, and 0
otherwise. The goal of the search, in our case, is to find a chromosome that
represents a set of features that lead to highest accuracy. If the result of
the search is the same for multiple feature subsets and all have the best accuracy, the one with the smallest cardinality is the desired one.\\
The result of the GA search gives the highest accuracy for 6 and 9 feature
subsets. But, the smallest chromosome string is selected as the desired
feature subset. The selected feature subset is given in Table2 and
clearly shows that the features related to the T wave is ignored. The medical
reason behind this fact is that, in the diagnosis of arrhythmias under our
consideration the T wave has no impact and usually is not used by practitioners. The parameters which are important for the diagnosis of arrhythmias under investigation are; depolarization of atrium and ventricle, and the relation between these two depolarizations. In other words, as repolarization has no role in heart rhythm detection so the T wave which is the product of repolarization will also lose its importance for our diagnosis.
\subsubsection{ Working principle}
GA begins with a set of solution and along with the ECG beat is given to the
ANN and the output of the ANN is compared with the desired output and hence the accuracy is calculated. If the accuracy is below the accepted level, the feedback loop is given to the GA block, to select the new population and the process repeats until the output error is minimized and the highest accuracy is achieved.  Figure \ref{GAfinal}-a shows the error plot of the feature subset selection process and Figure \ref{GAfinal}-b shows a block diagram of the system.

\begin{figure}[ht!]
\centering
\includegraphics[width=16cm]{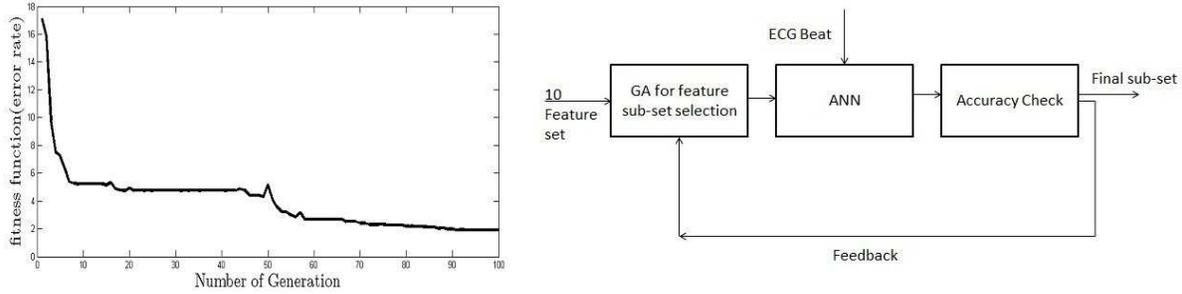}
\caption{(a) the fitness and (b) block diagram of the system}
\label{GAfinal}
\end{figure}

\subsubsection{Parameter extraction}
 The ECG feature extraction plays a very important role in disease diagnosis. The amplitudes and interval values of P-QRS-T waves determine the functionality of the heart of every human and the majority of the clinically important information in the ECG is in the intervals and amplitudes of these waves.
An accurate diagnosis of arrhythmia will be possible by the precise detection
of P, QRS and T waves. If the beat is similar to a normal beat it is easy to
detect the waves by eye, but it could be rather challenging in cases of an abnormal shape of the beat. The waves are studied by some parameters of amplitude and time. Some techniques have been proposed for automatic detection of these parameters \cite{Pantompkins,Sahoo}. In this work we use the wavelet transform technique. The wavelet transform is a transformation between time and frequency domains. In contrast to some transformation like Fourier transform, wavelet could be localized both in time and frequency\cite{Waveletdenoise}. The main equation of the transformation is:

\[ T(a,b)=\frac{1}{\sqrt{a}} \int _{-\infty} ^{ \infty} x(t)\psi^* ( \frac{t-b}{a})dt \]

$\psi$ is called the ‘mother wavelet’ and has to satisfy some
conditions\cite{Addison}. Parameters \textit{a} and \textit{b} are the width
of the signal and the time localization center respectively. In 1989
R. Coifman introduced a new class of wavelets which are called coiflet
\cite{Dau}. In this class of wavelet the mother wavelet must have zero
momentum. Another parameter in coiflet transform is order of coiflet. The
order of coiflet is chosen in a way that the transformed data best fits the
original data. The numerical procedures are done in Matlab using wavelet
toolbox and  the coiflet transform of order 1 is used \cite{Coiflet1}. Fig
\ref{QRSestimation} shows a transformed beat.

\begin{figure}[ht!]
\centering
\includegraphics[width=16cm]{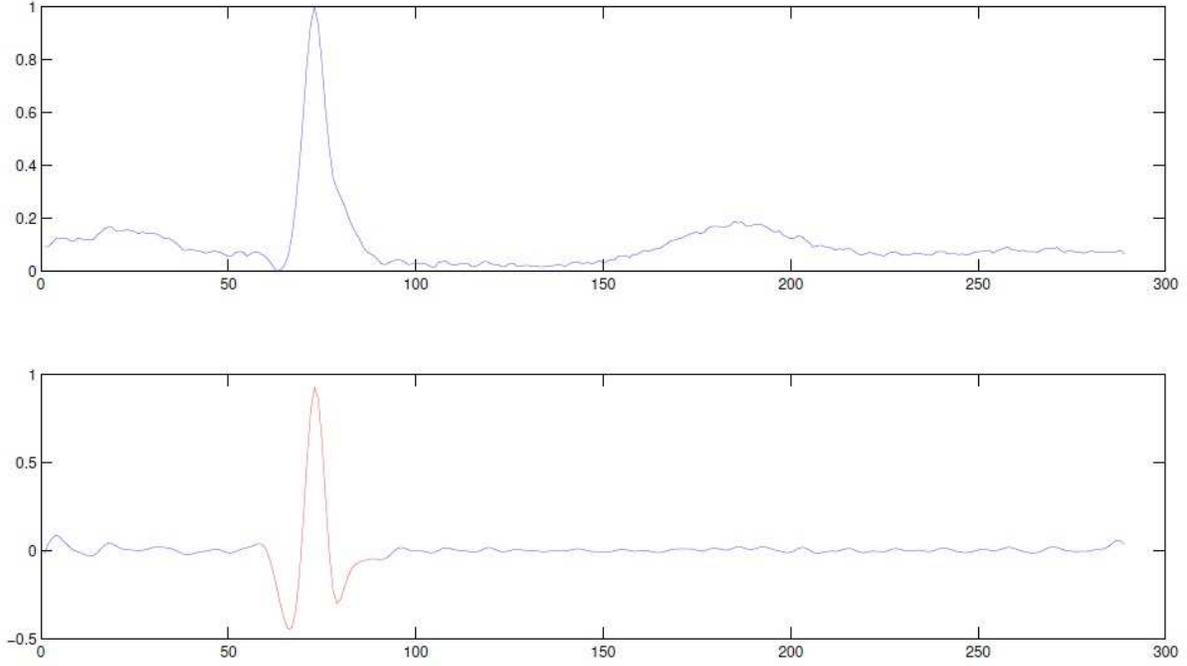}
\caption{QRS detection by coiflet transform of order 1.}
\label{QRSestimation}
\end{figure}

Using this transformation, Q, R, and S are determined; and knowing the approximate distance of P from these points, P is determined. 
Using this procedure the values in Table2 are obtained which are known as
morphological parameters. These parameters and the ECG beat will serve as the final input to the neural network. 
\\
Using wavelet transform, the small frequencies present in the signal are removed. So wavelet transform could be considered as a method to remove noise from the ECG signal as well.

\subsection*{Neural networks and classification}
We have used Probabilistic Neural Networks (PNN) among the ANN algorithms for
classification of input data. This algorithm learns the prediction of
Probability Density Function (PDF). It’s architecture consist of four layers, shown in Figure \ref{PNN}. 
The first layer is the input layer and fully connected to the next layer. Input vector $X = (x_1, x_2, ..., x_n ) \in \Re^n$ is applied to \textit{n} neurons of input layer. The second layer in which the input vectors are stored, is called the pattern layer. In the second layer including n*K neurons, the distance between the input vector and each one of K training examples is calculated; The third layer is the summation layer, which has K elements. Each element in this layer combines via processing elements in the pattern layer through the following estimator:
\begin{equation}
S_k (X) = \frac{1}{(2 \pi \sigma^2)^{n/2}}\frac{1}{N_k}\sum_{i=1}^{N_k}\exp \left( -
\frac{\parallel X - X_{k,i} \parallel^2}{2 \sigma^2} \right).
\end{equation}
where $\sigma$, known as the spread or smoothing parameter is the deviation of the Gaussian function and $ X_{k,i} \in  \Re ^n $ is the center of the kernel.
 Finally, the output layer selects the neuron in the summation layer with the maximum output\cite{Spetch}.
\begin{equation}
C(X)=\rm arg \ max\it(_{\substack{\\ \\ \hskip -0.8 cm 1 \le k \le
K}} S_k).
\end{equation}

\begin{figure}[ht!]
\centering
\includegraphics[width=16cm]{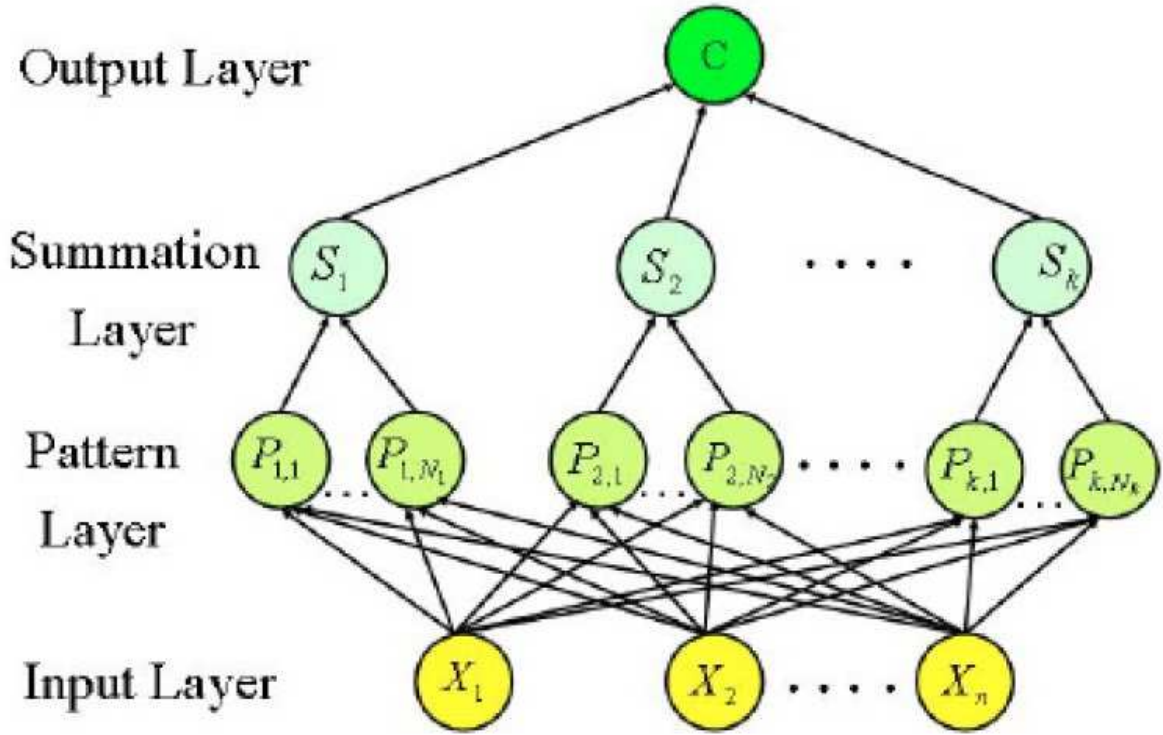}
\caption{Structure of the Probabilistic Neural Network.}
\label{PNN}
\end{figure}

\section*{Results and Discussion}

We have developed an automated system that works with the Probabilistic Neural
Network algorithm and gives a promising result; which we evaluated based on Sensitivity, Specificity, and Accurcy which are defined as the following.

\[ Sensitivity=\frac{TP}{TP+FN} \]
\[ Specificity=\frac{TN}{TN+FP} \]
\[ Accuracy=\frac{TN+TP}{TN+TP+FN+FP} \]

One can find the definitions for TP,TN,FP and FN in Table4.  
Depending on the input type we had two approaches to deal with.

\subsection*{First approach} 
The simplest way is to give the whole information of a beat (the
voltage-time), that contains 289 data points, as an input to the ANN. Before
giving it to the network, we applied normalization and de-noising on the data and the result is given in Table3-1. 
As it is shown in the table we used different ratios of beats for training and
testing the PNN. For example in feature set I, 250 beats out of 350 are
selected for the training set and the rest kept as the testing set. Similarly,
the ratio of 300 to 50 for feature set II, ratio of 150 to 200 for feature
set III and finally the ratio is 200 to 150 for feature set IV.
\subsection*{Second approach} 
In this scheme all the preprocessing steps are taken and the final reduced ECG
beat  contains 29 data points. We have also extracted the morphological
parameters of each beat and then the optimized feature subset is chosen.  The
addition of the ECG beat and the selected morphological parameters will form
our input vector to the neural network with 35 data points. The result of this
method is given in Table3-2.  The number of train and test sets for Table3-2
is the same as the previous section.




As it appears in Table3-1 and Table3-2 the results obtained by the second
approach has a higher accuracy.
\\
In this study we are working on 8 types of arrhythmia whereas previous
attempts are made on 6 types. By increasing the number of arrhythmias under
consideration the risk of misclassification can increase. But, the highest classification accuracy achieved by the second approach is \%99.42 which we claim is the highest accuracy for 8 types of arrhythmia ever made.

\section*{Conclusions}
  In this paper ANN is applied for the classification of ECG beat arrhythmias. The results show a high rate of accuracy in 8 classes of arrhythmias using probabilistic neural networks(PNN). 
The selection of ECG beats used in training set plays an important role.
This set has to contain combination of different arrhythmias and all their possible variations from different patients to achieve good results.
Two types of input vectors to the ANN are considered. First the whole ECG signal is taken as input; in  which case the results are good. 
In the second input type the reduced data along with the important parameters of ECG beat is considered as input; this way  the ANN has better performance because of the reduced overhead. 
The time and size of the network is also reduced and the accuracy acheived this way is higher than the first input data type.
Wavelet transform of type coilflet is used to extract the parameters from ECG
waves. It is the most efficient compared to other methods.
The set of parameters fed to the network is also very important. To choose the optimum set of parameters among all parameters 
Genetic Algorithm is used. GA selects the best parameters without affecting
the performance of ANN. The set of parameters chosen by GA is in good
agreement with the set suggested by physicians to detect arrhythmia type.\\
Similar methods could be used to study the anatomical disease and also
classification of EEG signals. We are planing to sudy the brain related
disease in future, using MRI images and the EEG signal.


\section*{Acknowledgements}
  \ifthenelse{\boolean{publ}}{\small}{}
  
  This study was financially supported by Iran National Science
  Foundation(Project No 90000947); and also the authors would like to thank
  Prof. Y. Sobouti and Dr. A. Biglari for initiating this project and
  establishing the collaboration with Zanjan University of Medical Sciences.


{\ifthenelse{\boolean{publ}}{\footnotesize}{\small}
 \bibliographystyle{bmc_article}  
  \bibliography{ECG_article} }     


\ifthenelse{\boolean{publ}}{\end{multicols}}{}



\section*{Figures}
  \subsection*{Figure 1 - Sample Normal Beat}
	The sample normal beat showing different morphological parameters.
  \subsection*{Figure 2 - Sample signals for eight arrhythmias}
      	Sample signals for eight arrhythmias under consideration.
  \subsection*{Figure 3 - Filtered signal with median algorithm}
	Median algorithm used for data reduction and 10 consecutive points are
        replaced by a single point with their mean value.
  \subsection*{Figure 4 - Important morphological parameter selection }
	(a) The fitness and (b) block diagram of the system
  \subsection*{Figure 5 - The QRS estimation}
	The result of Wavelet transform (Coiflet1) on sample ECG signal for
        the QRS estimation
   \subsection*{Figure 6 - The PNN architecture}
	The architecture of the Probabilistic Neural Network



\section*{Tables}

\subsection*{Table 1 - Types of arrhythmia}
The eight types of arrhythmia
\begin{table}[h!b!p!]
\label{Table1}
\begin{tabular}{|c|c|}
\hline symbol &  arrythmia \\ 
\hline N &  Normal beat\\ 
\hline / &  Paced beat \\ 
\hline A &  Premature beat (atrial,aberrated atrial,nodal) \\ 
\hline E &  Escape  beat(ventricular,atrial,nodal) \\ 
\hline f &  Fusion of paced and normal beat\\ 
\hline F &  Fusion of ventricular and normal beat\\ 
\hline L &  Left bundle branch block beat\\ 
\hline R &  Right bundle branch block beat\\ 
\hline 
\end{tabular} 
\end{table}
\vspace{30cm}

  \subsection*{Table 2 - ECG Parameters}
The ECG optimized morphological parameters using Genetic Algorithm.
\begin{table}[h!b!p!]
\label{Table2}
\begin{tabular}{|c|c|}
\hline Feature & Duration \\ 
\hline P wave  & 60-80(ms) \\ 
\hline PR-segment & 50-120(ms) \\ 
\hline PR- interval  & 120-200(ms) \\ 
\hline QRS duration  & 80-120(ms) \\ 
\hline QRS amplitude  & 0.25-1(v) \\ 
\hline RR-interval  & 400-1200(ms) \\ 
\hline 
\end{tabular} 
\end{table}

  \subsection*{Table 3-1 - The result of first approach}
The results of first approach for different ratios of training and testing data sets as input to the PNN. \\
    \par
    \mbox{
\begin{tiny}
\begin{tabular}{|c|c|c|c|c|c|c|c|c|c|c|c|c|}
\hline ECG & \multicolumn{3}{|c|}{Feature Set I} &  \multicolumn{3}{|c|}{Feature Set II}  &  \multicolumn{3}{|c|}{Feature Set III}  &  \multicolumn{3}{|c|}{Feature Set IV} \\ \cline{2-13}
beat type & Sens(\%)  &Spec(\%)  &Acc(\%)  &Sens(\%)  &Spec(\%)  &Acc(\%)  &Sens(\%)  &Spec(\%)  &Acc(\%)  &Sens(\%)  &Spec(\%)  &Acc(\%)  \\   
\hline N & 95.00 &99.52  &98.79  &90.36  &98.30  &97.11  &89.71  &97.89  &96.67  &93.00  &99.35  &98.39  \\ 
\hline R &100.0  &90.53  &91.94  &100.0  &99.49  &99.56  &100.0  &94.29  &95.13  &100.0  &96.51  &96.89  \\ 
\hline L &100.0  &99.51  &99.59  &98.00  &100.0  &99.69  &97.14  &97.96  &97.84  &100.0  &99.10  &99.24  \\ 
\hline F &83.50  &91.03  &89.93  &77.82  &99.37  &96.19  &81.71  &99.19  &96.59  &88.33  &91.93  &91.43  \\ 
\hline f &67.50  &96.82  &92.42  &78.81  &98.41  &95.67  &74.85  &96.25  &93.17  &75.00  &97.80  &94.51  \\ 
\hline E &39.00  &96.73  &88.45  &77.83  &94.43  &92.50  &53.71  &95.76  &89.92  &59.33  &95.50  &90.50  \\ 
\hline A &32.00  &97.74  &88.32  &71.24  &96.35  &94.11  &50.85  &96.22  &89.92  &39.33  &98.66  &90.46  \\ 
\hline / &99.50  &96.09  &96.63  &100.0  &97.64  &97.99  &100.0  &97.36  &97.75  &99.66  &97.08  &97.46  \\ 
\hline Average &77.06  &95.99  &93.25 &86.75  &97.99  &93.22  &80.99  &96.86  &94.62  &81.83  &96.99  &94.86  \\ 
\hline 
\end{tabular} 
\end{tiny}
      }
  \subsection*{Table 3-2 - The result of second approach}
    The results of second approach for different ratios of
    training and testing data sets as input to the PNN. 
\begin{tiny}
\begin{tabular}{|c|c|c|c|c|c|c|c|c|c|c|c|c|}
\hline ECG & \multicolumn{3}{|c|}{Feature Set I} &  \multicolumn{3}{|c|}{Feature Set II}  &  \multicolumn{3}{|c|}{Feature Set III}  &  \multicolumn{3}{|c|}{Feature Set IV} \\ \cline{2-13}
beat type & Sens(\%)  &Spec(\%)  &Acc(\%)  &Sens(\%)  &Spec(\%)  &Acc(\%)  &Sens(\%)  &Spec(\%)  &Acc(\%)  &Sens(\%)  &Spec(\%)  &Acc(\%)  \\ 
\hline N          &99.00    &100.0   &99.72    &100.00   &99.41    &99.48    &100.00   &99.70    &99.74    &100.00    &99.57    &99.63  \\ 
\hline R          &80.00    &94.35   &95.08    &100.00   &99.70    &99.74    &94.00     &98.99    &98.36    &100.00    &98.98    &99.11 \\ 
\hline L           &88.00    &100.0   &98.39    &100.00   &100.00  &100.00  &99.00     &100.00  &99.87    &99.20    &99.69    &99.63  \\ 
\hline F           &96.00   &98.91    &98.52    &96.00     &99.70    &99.23    &99.00     &99.85    &99.74    &88.00   &98.64     &97.28  \\ 
\hline f            &96.00   &99.84    &99.32    &100.00   &98.82   &98.87    &100.00    &99.85    &99.87    &98.40    &99.22    &99.11 \\ 
\hline E           &93.00   &99.22    &98.39    &90.00     &98.84    &97.72    &95.00      &100.00  &99.36    &84.40       &99.46       &97.68  \\ 
\hline A           &63.00   &98.10    &93.63    &88.00     &100.00  &98.47   &96.00     &99.13     &98.73    &90.40    &99.28    &98.14  \\ 
\hline /            &100.0   &100.0    &100.0    &100.0     &99.70    &99.74   &99.00     &99.85    &99.74    &99.60    &98.98    &99.06  \\ 
\hline Average&91.97   &98.80    &97.88    &96.75     &99.52    &99.16   & 97.75    &99.67    &99.42    &95.00    &99.22    &98.71  \\ 
\hline 
\end{tabular} 
\end{tiny}

  \subsection*{Table 4 - Definitions of TP, TN, FP, FN}
    Definition of TP, TN, FP, FN
\begin{table}
\label{Table4}
\begin{tabular}{|c|c|c|c|}
\hline
\multicolumn{2}{|c|}{} & \multicolumn{2}{|c|} {Reality}   \\ \cline{3-4}
\multicolumn{2}{|c|}{}  & Sample is  A type. & Sample is not A type. \\ \cline{2-4}
\hline Outcome of network weather & True & True Poseitive(TP) &  False Positive(FP)\\ \cline{2-4}
the sample is A type. & False & False Negative(FN) & True Negative(TN) \\ 
\hline 
\end{tabular} 
\end{table}




\end{bmcformat}
\end{document}